\documentclass{article}
\usepackage{spconf,amsmath,graphicx,hyperref}

\usepackage{hyperref}
\usepackage{url}

\usepackage{amssymb}
\usepackage{times}
\usepackage{latexsym}
\usepackage{amsmath}
\usepackage{amssymb}
\usepackage{enumitem}
\usepackage{graphicx}
\usepackage{pifont}
\usepackage{color}
\usepackage{titletoc}
\usepackage{colortbl}
\usepackage{wrapfig}  
\usepackage{lipsum} 
\usepackage{multirow}
\usepackage{listings}
\usepackage{array}

\usepackage{booktabs} 
\usepackage{caption}  

\usepackage{inconsolata}

\usepackage{array}
\usepackage{amsfonts}
\usepackage[ruled,linesnumbered,noend]{algorithm2e}
\usepackage{algpseudocode}
\usepackage{fontawesome}

\usepackage{fontawesome}

\title{ 
Making Medical Vision-Language Models Think Causally 
\\
Across Modalities with Retrieval-Augmented Cross-Modal  Reasoning
}

\name{
Weiqin Yang$^{1,*}$ \quad
Haowen Xue$^{2,*,\dagger}$\thanks{$^*$~Equal contribution. $\dagger$~Corresponding author.}
 \quad
Qingyi Peng$^{3}$ \quad
Hexuan Hu$^{2}$ \quad
Qian Huang$^{2}$ \quad
Tingbo Zhang$^{2}$
}

\address{
$^{1}$University of Adelaide  \quad  $^{2}$Hohai University \quad
$^{3}$Amap \\
}

\usepackage{xcolor}
\definecolor{mildteal}{RGB}{0, 102, 102}

\begin{document}
%
\maketitle
%

\newcommand{\our}{\textsc{MCRAG}}

\begin{abstract}

Medical vision–language models (VLMs) achieve strong performance in diagnostic reporting and image–text alignment, yet their underlying reasoning mechanisms remain fundamentally correlational, exhibiting reliance on superficial statistical associations that fail to capture the causal pathophysiological mechanisms central to clinical decision-making. This limitation makes them fragile, prone to hallucinations, and sensitive to dataset biases. Retrieval-augmented generation (RAG) offers a partial remedy by grounding predictions in external knowledge. However, conventional RAG depends on semantic similarity, introducing new spurious correlations. We propose  Multimodal Causal Retrieval-Augmented Generation (\our), a framework that integrates causal inference principles with multimodal retrieval. \our~retrieves clinically relevant exemplars and causal graphs from external sources, conditioning model reasoning on counterfactual and interventional evidence rather than correlations alone. Applied to radiology report generation, diagnosis prediction, and visual question answering, \our~improves factual accuracy, robustness to distribution shifts, and interpretability. Our results highlight causal retrieval as a scalable path toward medical VLMs that think beyond pattern matching, enabling trustworthy multimodal reasoning in high-stakes clinical settings.





\end{abstract}


\section{Introduction}
\label{sec:introduction}

Artificial Intelligence (AI) has already transformed healthcare and continues to hold substantial potential for further innovation within clinical ecosystems. 
Recently, Medical Large Vision-Language Models (Med-LVLMs) have shown great promise for advancing interactive and intelligent diagnosis~\cite{li2023llava,wu2023towards}. 
Despite this potential, 
current Med-LVLMs still face significant reliability issues, particularly their tendency to generate non-factual medical 
responses~\cite{chen2024detecting}, making them unreliable in critical medical applications. These factuality issues raise serious concerns when deploying such models in clinical settings, where even small diagnostic errors could lead to severe consequences for patient care.

Recently, researchers have begun to focus on improving the factuality of Med-LVLMs through various techniques, including fine-tuning~\cite{li2023llava}, low-rank adaptation~\cite{hu2022lora}, and retrieval-augmented generation (RAG)~\cite{gao2023retrieval,xiong2024benchmarking}. Fine-tuning is a direct method to improve model performance, but faces several limitations in the medical field. First, there is a lack of sufficient high-quality labeled data for fine-tuning in the medical domain. Second, a distribution shift often exists between training datasets the real-world deployment data, leading to significantly worse model performance during deployment. Hence, Retrieval-Augmented Generation (RAG)~\cite{gao2023retrieval} has emerged as a promising solution, grounding model outputs in external knowledge to improve factuality. Recent works adapted RAG for medicine, including MedRAG~\cite{xiong2024benchmarking}, MMed-RAG~\cite{xia2024mmed}. 
Despite improvements, current methods remain vulnerable to semantic over-reliance, cross-modality misalignment, and spurious correlations, largely due to mismatch between retrieved contexts and visual–language grounding.

 



Causality-based methods seek to improve retrieval accuracy and representation learning, but critical limitations persist~\cite{chen2024domain}. CausalRAG~\cite{wang2025causalrag} ranks contexts by causal importance, but does not address cross-modal alignment. Similarly, CMCRL~\cite{chen2025cross} learns shared causal representations but underuses the structural dependencies needed for medical reasoning. Although causal graphs improve interpretability~\cite{pearl2009causality,ma_causal_2024}, current approaches mainly rely on language-only causal discovery~\cite{wang2025causalrag}. 

These limitations necessitate a more holistic framework that simultaneously addresses factuality and alignment by integrating causal reasoning with a multimodal structure. Accordingly, we propose to build explicit causal graphs from multimodal data and use them to guide RAG retrieval for medical reasoning.

\begin{figure*}
    \centering
    \includegraphics[width=1\linewidth]{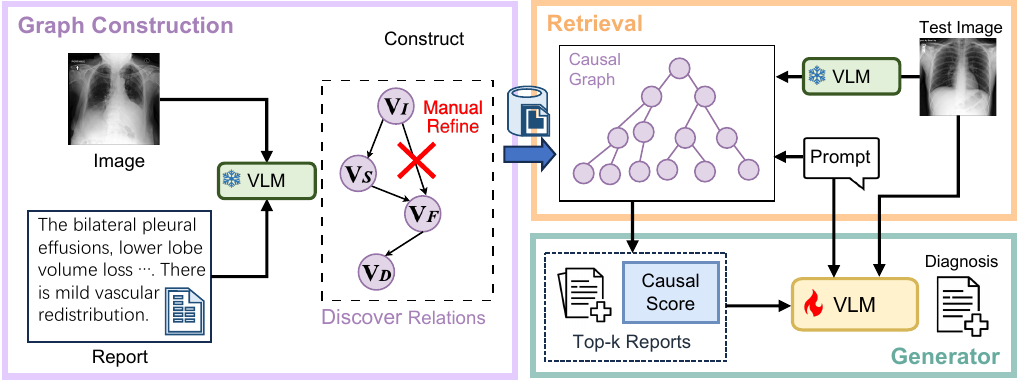}
    \caption{\textbf{MCRAG overview.}   Left (Graph Construction): A VLM extracts entities and relations from paired images and reports to construct a causal graph, followed by manual refinement to prune spurious links. Right (Retrieval and Generation): For a test image, the VLM queries the causal graph to retrieve top-k relevant reports ranked by a causal score. The retrieved reports and test image are then combined into a prompt for the generator VLM, which produces the final diagnosis. }
    \label{fig:placeholder}
\end{figure*}

In this paper, we propose \our~(Multimodal Causal Retrieval-Augmented Generation), a retrieval framework designed to improve factuality and robustness in Med-LVLMs.  \our~introduces a causal alignment graph constructed from verifiable medical literature, capturing structured dependencies between visual and textual modalities. This graph enables retrieval guided not only by semantic similarity but also by causal and structural relevance, thus mitigating spurious correlations. Furthermore, \our~incorporates RAG-based preference fine-tuning to enforce two key principles: (i) grounding responses in input images when relevant to prevent degenerate text-only outputs; and (ii) interpreting retrieved contexts causally to enhance robustness under uncertainty. Finally, \our~applies causal filtering to balance coverage and precision in retrieval, selecting context based on structural importance rather than raw similarity.

\textbf{Our contributions are threefold.}
\begin{itemize}
  \item \textbf{\our:} We introduce the first framework that integrates causal graphs with cross-modal alignment for medical vision–language generation.
  \item \textbf{Causal alignment graph:} We design a knowledge-guided graph that enables structured cross-modal retrieval grounded in medical semantics.
  \item \textbf{Preference fine-tuning:} We propose a strategy that enforces image-grounded, causally coherent generation, improving robustness and factual accuracy across medical tasks.
\end{itemize}

\vspace{-0.3em}
\section{Methodology}
\vspace{-0.3em}

In this section, we present \textbf{MCRAG}—a multi-modal, causal retrieval-augmented generation framework that improves the factuality of Med-LVLMs by tightly coupling retrieval with an explicit Structural Causal Model (SCM). The framework has three stages: (1) \textit{domain-aware retrieval}, which selects the optimal retriever for each input; (2) \textit{adaptive context selection}, which filters and sizes evidence on the fly; and (3) \textit{RAG-based preference fine-tuning}, which aligns responses with SCM-supported evidence. 




\subsection{Structural Causal Model (SCM)}
At the core of the MCRAG framework is the formal representation of medical knowledge as a \textit{Structural Causal Model} (SCM), a mathematical formalism for causal inference. An SCM, denoted as \(\mathcal{M}\), is a tuple \(\langle \mathbf{V}, \mathbf{U}, \mathcal{F} \rangle\), where:

\begin{itemize}
    \item \(\mathbf{V}\) is a set of endogenous variables, representing the manifest, observable variables within the system. 
    In the medical context, \(\mathbf{V}\) includes variables corresponding to image regions (\(V_I\)), 
    clinical findings (\(V_F\)), patient symptoms (\(V_S\)), and diagnostic outcomes (\(V_D\)).

    \item \(\mathbf{U}\) is a set of exogenous variables, representing latent or unobserved factors.  These variables account for all factors influencing the endogenous variables that are not explicitly included in the model, 
    such as genetic predispositions or data heterogeneity across hospital systems.

    \item \(\mathcal{F}\) is a set of structural equations, one for each variable \(V_i \in \mathbf{V}\). 
    Each equation \(f_i \in \mathcal{F}\) defines the value of \(V_i\) as a function of its parents, 
    \(\mathrm{Pa}(V_i)\), in the causal graph and its corresponding exogenous variable \(U_i \in \mathbf{U}\). 
    For instance, a function for a specific clinical finding might be expressed as: 
    \begin{equation}
        \hat V_F = f_F(\mathrm{Pa}(V_F), U_F)
    \end{equation}
        
\end{itemize}

A key feature of the SCM is its associated causal graph \(G\), a directed graph over the variables in \(\mathbf{V}\) and \(\mathbf{U}\).

\subsection{Cross-Modal Medical Causal Graph Construction}

The construction of a comprehensive causal graph \(G\) requires integrating information from multiple modalities. Our framework employs a two-stage data-driven causal discovery protocol to build \(G\) from a corpus of paired medical images and clinical reports. \\

\noindent\textbf{Step 1: Multimodal VLMs-Assisted Causal Discovery.} 
We use vision-Language Models (VLMs) to serve as the primary knowledge extractor. The model is prompted to analyze image-text pairs to identify potential causal relationships. For instance, a visual feature like  \textbf{‘pulmonary opacity’} observed in a chest X-ray would be linked to the textual entity \textbf{‘pneumonia’} in the accompanying report, proposing a causal edge between them, grounding textual concepts in visual evidence. 
Let $v_i$ and $r_j$ denote the visual embedding of image $i$ and textual  report embedding of report $j$, respectively. The retriever’s contrastive loss $\mathcal{L}{\mathrm{retr}}$ maximizes the cosine similarity $s_{ij}= \langle v_i, r_j\rangle$ for true image–report pairs while minimizing $s_{ij}$ for mismatched pairs. In practice, we collect a corpus of domain-specific text (e.g. reports for radiology images) and use these as the knowledge base. \\

\noindent\textbf{Step 2: Manual Graph Refinement.} 
Starting from the draft graph proposed by the VLMs under a low-confidence threshold, we conduct a principled manual review of every candidate causal edge. Each edge is evaluated for clinical plausibility and statistical support (e.g., whether the conditional probability of a diagnosis given a visual feature corresponds with domain knowledge), and any edge failing this inspection is removed. For instance, if a visual feature $V_I$ and the final diagnosis $V_D$ are conditionally independent given a textual clinical finding $V_F$ (i.e., $V_I \perp\!\!\!\perp V_D \mid V_F$), this provides statistical evidence for the causal pathway $V_I \rightarrow V_F \rightarrow V_D$ and justifies pruning the spurious direct edge $V_I \rightarrow V_D$. Clinically unreasonable edges are discarded even if strong statistical associations appear.

\subsection{Causal-based Retrieval Augmented Reasoning}

Given an input image \(I\), we first retrieve the top-\(K\) nearest textual reports \(\{R_k\}_{k=1}^{K}\) in a joint embedding space. We then enforce causal consistency using the graph \(G\). For each candidate \(R_k\), we extract the variables it references (e.g., findings \(V_F\) and diagnoses \(V_D\)) and evaluate how well they are supported by image-derived features \(V_I\) along the causal paths in \(G\) (preferably \(V_I \rightarrow V_F \rightarrow V_D\)).

\begin{equation}
\begin{aligned}
\operatorname{Score}(R_k)
&= (1-\alpha)\,\log p_G\!\big(V_D,V_F \mid V_I\big)\\
&\quad {}+ \alpha\,\mathrm{sim}(I,R_k),\qquad \alpha\in[0,1]
\end{aligned}
\end{equation}

\noindent\textit{where} $\mathrm{sim}(I,R_k)$ denotes the image–report embedding similarity, and $p_G(\cdot)$ is the likelihood induced by the factorization implied by the causal graph $G$. For example, if $G$ retains the mediated path $V_I \to V_F \to V_D$, then
\begin{equation}
    p_G(V_D,V_F \mid V_I)=p(V_F \mid V_I)\,p(V_D \mid V_F)
\end{equation}

Candidates consistent with $G$ are up-weighted, whereas those relying on unsupported or pruned edges are down-weighted or discarded, yielding retrieved reports that are both semantically relevant and causally grounded.

After assembling high-quality retrieved contexts and their associated causal relations, MCRAG integrates them into the generation process via retrieval-augmented fine-tuning.

\vspace{-1em}
\section{Experiment}

\begin{table*}[t!]
    \centering
    \footnotesize
    \caption{Performance (\%) of different methods on Radiology VQA and Radiology Report Generation. For VQA, we report Accuracy, F1 score, and AUROC; for Report Generation, we report BLEU, ROUGE-L (R-L), and METEOR (MET). The best and second-best results are highlighted in \colorbox{red!15}{red} and \colorbox{blue!15}{blue}, respectively. Comparison results are reported from MMed-RAG~\cite{xia2024mmed}.} 
    \vspace{-1em}
    \resizebox{\linewidth}{!}{
    \begin{tabular}{l||ccc|ccc||ccc|ccc}
    \toprule
    \multirow{2}{*}{Models} 
    & \multicolumn{6}{c||}{\textbf{Radiology VQA}} & \multicolumn{6}{c}{\textbf{Radiology Report Generation}} \\
    \cmidrule(r){2-7} \cmidrule(r){8-13}
    & \multicolumn{3}{c|}{IU-Xray} & \multicolumn{3}{c||}{MIMIC-CXR} & \multicolumn{3}{c|}{IU-Xray} & \multicolumn{3}{c}{MIMIC-CXR} \\
    \cmidrule(r){2-4} \cmidrule(r){5-7} \cmidrule(r){8-10} \cmidrule(r){11-13}
    & Acc~$\uparrow$ & F1~$\uparrow$ & AUC~$\uparrow$
    & Acc~$\uparrow$ & F1~$\uparrow$ & AUC~$\uparrow$
    & BLEU~$\uparrow$ & R-L~$\uparrow$ & MET~$\uparrow$
    & BLEU~$\uparrow$ & R-L~$\uparrow$ & MET~$\uparrow$ \\

    \midrule
LLaVA-Med-1.5~\cite{li2023llava} 
& 75.47 & 64.04 & 67.46 & 75.79 & 80.49 & 68.84 & 9.64 & 12.26 & 8.21 & 12.11 & 13.05 & 11.16 \\
\midrule
+ DoLa~\cite{chuang2023dola} 
& 78.00 & 66.75 & 72.19 & 81.35 & 85.73 & 72.73 & 11.79 & 15.82 & 12.72 & 17.11 & 14.89 & 14.81 \\
+ OPERA~\cite{huang2023opera} 
& 70.59 & 61.54 & 63.22 & 69.34 & 76.66 & 62.46 & 10.66 & 14.70 & 12.01 & 15.40 & 12.52 & 13.72 \\
+ VCD~\cite{leng2023mitigating} 
& 68.99 & 54.35 & 61.08 & 70.89 & 75.57 & 64.61 & 10.42 & 14.14 & 11.59 & 15.18 & 12.30 & 13.38 \\
\midrule
+ MedDr~\cite{he2024meddr} 
& 83.33 & 67.80 & 77.15 & 55.16 & 56.18 & 58.47 & 12.37 & 16.45 & 13.50 & {18.59} & 15.72 & 16.77 \\
+ FactMM-RAG~\cite{sun2024fact} 
& 84.51 & 68.51 & 77.07 & 77.58 & 81.86 & 70.09 & 14.70 & 18.05 & 15.92 & 18.71 & 15.84 & 16.82 \\
+ RULE~\cite{xia2024rule} 
& 87.84 & 78.00 & 85.78 & \cellcolor{blue!15}{83.92} & 87.49 & 83.44 & 27.53 & 23.16 & 27.99 & 18.61 & \cellcolor{blue!15}{15.96} & {17.42} \\
+ MMed-RAG~\cite{xia2024mmed} 
& \cellcolor{blue!15}{89.54} & \cellcolor{blue!15}{80.72} & \cellcolor{blue!15}{87.13} & 83.57 & \cellcolor{blue!15}{88.49} & \cellcolor{blue!15}{85.08} & \cellcolor{blue!15}{31.38} & \cellcolor{blue!15}{25.59} & \cellcolor{blue!15}{32.43} & \cellcolor{blue!15}{23.25} & 12.34 & \cellcolor{blue!15}{20.47} \\
\midrule
+ \our 
& \cellcolor{red!15}{90.12} & \cellcolor{red!15}{82.03} & \cellcolor{red!15}{88.25} 
& \cellcolor{red!15}{84.91} & \cellcolor{red!15}{89.37} & \cellcolor{red!15}{86.42} 
& \cellcolor{red!15}{35.02} & \cellcolor{red!15}{28.47} & \cellcolor{red!15}{35.18} 
& \cellcolor{red!15}{25.81} & \cellcolor{red!15}{15.05} & \cellcolor{red!15}{22.34} \\

    \bottomrule
    \end{tabular}
    }
    \vspace{-1em}
    \label{tab:radiology}
\end{table*}

\subsection{Experimental Setups}
\label{sec:exp}

 For the language model, we adopt LLaVA-Med-1.5-7B~\cite{li2023llava}, fine-tuned with LoRA~\cite{hu2021lora} using the AdamW optimizer. The fine-tuning is performed with a learning rate of $3\times10^{-5}$, weight decay of $10^{-2}$, a batch size of 16, and for 500 epochs. For modality-specific encoders, we employ MedVIT~\cite{manzari2023medvit} as the vision encoder and BioClinicalBERT~\cite{alsentzer2019publicly} as the text encoder.

We adopt the experimental framework of MMed-RAG\cite{xia2024mmed} and evaluate hallucination mitigation methods from two complementary perspectives. Decoding-based approaches, such as  DoLa~\cite{chuang2023dola}, OPERA~\cite{huang2023opera}, and VCD~\cite{leng2023mitigating}, improve factual consistency by directly adjusting the model’s output distribution. In contrast, multimodal retrieval-augmented generation (RAG) methods, including MedDr~\cite{he2024meddr}, FactMM-RAG~\cite{sun2024fact}, RULE~\cite{xia2024rule}, and MMed-RAG~\cite{xia2024mmed}, mitigate hallucinations by grounding responses in external knowledge.  We didn't choose the CasualRAG is because it is not multi-modal, so not in our scope.

Our experiments employ MIMIC-CXR~\cite{johnson2019mimic} and IU-Xray~\cite{demner2016preparing} as benchmark datasets. Question–answer pairs are taken from MMed-RAG~\cite{xia2024mmed}. Following prior work~\cite{xia2024mmed}, we assess medical VQA performance using Accuracy, F1 Score, and AUROC, while report generation is evaluated with BLEU, ROUGE-L, and METEOR.


\vspace{-1em}
\subsection{Comparison Results}

Table~\ref{tab:radiology} compares decoding-only baselines with retrieval-augmented models. While MMed-RAG delivers strong results (e.g., $89.54$ Acc and $87.13$ AUC on IU-Xray VQA), our method (\our) consistently sets new state-of-the-art across all tasks. On IU-Xray VQA, \our surpasses MMed-RAG by $+0.58$ Acc, $+1.31$ F1, and $+1.12$ AUC; on MIMIC-CXR VQA, it achieves further gains of $+1.34$ Acc, $+0.88$ F1, and $+1.34$ AUC. For report generation, \our raises BLEU to $35.02$ and $25.81$, improving over MMed-RAG by $+3.64$ and $+2.56$ on IU-Xray and MIMIC-CXR, respectively. 
These results demonstrate that causality-guided retrieval not only enhances factual accuracy in VQA but also yields more fluent, faithful clinical reports.\\

\noindent
\textbf{Ablation Studies.} To understand the role of causality, we ablate both its presence and the ratio used for refining (i.e., the percentage of the causal branch manually removed). As shown in Table~\ref{tab:ablation}, removing causality causes the steepest drop ($-3.65$ Acc, $-3.95$ F1, $-3.34$ BLEU), highlighting its central role in grounding answers in clinically meaningful evidence. Using causality without refining partially recovers performance but still introduces noisy links. Introducing confidence-based refining steadily improves results, with the best trade-off observed at $\tau=0.7$ ($84.91$ Acc, $89.37$ F1, $25.81$ BLEU). Lower ratio (e.g., $\tau=0.5$) allow noise to persist, while higher ratio (e.g., $\tau=0.9$) over-prune and reduce recall. Causality thus drives robust reasoning by structuring the search space, while manually refining calibrates the precision–coverage trade-off by pruning unreliable links. 


Table~\ref{tab:rag_reports} shows that both re-ranking and filtering are crucial for RAG. Removing re-ranking reduces performance (83.78 Acc, 87.20 F1, 24.61 BLEU), while removing filtering leads to an even larger drop (82.15 / 85.40 / 23.20), indicating its stronger role. Varying $K$ reveals the evidence–noise trade-off: too few reports ($K=5$) limit coverage, too many ($K=20$) add noise, and the best balance is at $K=10$.

\begin{table}[t!]
\centering
\footnotesize
\caption{Ablation study of causality in \our~on the MIMIC-CXR dataset. 
We report Accuracy (Acc), F1, and BLEU as mean~$\pm$~std over 3 runs. 
$\tau$ denotes the ratio of causal branches removed for refining causal links.}
\label{tab:ablation}
\vspace{-0.5em}
\resizebox{0.48\textwidth}{!}{%
\begin{tabular}{l|ccc}
\toprule
Method & Acc~$\uparrow$ & F1~$\uparrow$ & BLEU~$\uparrow$ \\
\midrule
\our (Full Model, $\tau=0.7$) & \textbf{84.91 $\pm$ 0.21} & \textbf{89.37 $\pm$ 0.18} & \textbf{25.81 $\pm$ 0.42} \\
w/o Causality Relation        & 81.26 $\pm$ 0.33 & 86.71 $\pm$ 0.29 & 23.58 $\pm$ 0.55 \\
w/o Manual Refining           & 80.34 $\pm$ 0.27 & 85.42 $\pm$ 0.31 & 22.47 $\pm$ 0.61 \\
\midrule
$\tau=0.5$                    & 83.47 $\pm$ 0.24 & 87.92 $\pm$ 0.22 & 24.71 $\pm$ 0.48 \\
$\tau=0.7$ (ours)             & \textbf{84.91 $\pm$ 0.21} & \textbf{89.37 $\pm$ 0.18} & \textbf{25.81 $\pm$ 0.42} \\
$\tau=0.9$                    & 84.12 $\pm$ 0.26 & 88.41 $\pm$ 0.25 & 25.02 $\pm$ 0.47 \\
\bottomrule
\end{tabular}
}
\end{table}

\begin{table}[t!]
\centering
\footnotesize
\caption{Analysis of report usage in RAG with different retrieval settings. 
We report performance (mean~$\pm$~std over 3 runs). 
\emph{Re-ranking} reorders retrieved reports, and \emph{Filtering} removes low-quality ones by score threshold.}
\label{tab:rag_reports}
\vspace{-0.5em}
\resizebox{0.48\textwidth}{!}{%
\begin{tabular}{l|ccc}
\toprule
Method  & Acc~$\uparrow$ & F1~$\uparrow$ & BLEU~$\uparrow$ \\
\midrule
RAG (Full Model, $K=10$) & \textbf{84.91 $\pm$ 0.22} & \textbf{89.37 $\pm$ 0.20} & \textbf{25.81 $\pm$ 0.45} \\
w/o Re-ranking           & 83.78 $\pm$ 0.31 & 87.20 $\pm$ 0.27 & 24.61 $\pm$ 0.52 \\
w/o Filtering            & 82.15 $\pm$ 0.29 & 85.40 $\pm$ 0.33 & 23.20 $\pm$ 0.58 \\  
\midrule                  
$K=5$                    & 84.10 $\pm$ 0.25 & 87.40 $\pm$ 0.28 & 25.00 $\pm$ 0.49 \\
$K=10$ (ours)            & \textbf{84.91 $\pm$ 0.22} & \textbf{89.37 $\pm$ 0.20} & \textbf{25.81 $\pm$ 0.45} \\
$K=20$                   & 84.60 $\pm$ 0.27 & 88.00 $\pm$ 0.26 & 25.60 $\pm$ 0.50 \\
\bottomrule
\end{tabular}
}
\end{table}

\section{Limitations}
While \our~advances retrieval by incorporating causal reasoning, several limitations remain. The framework presupposes that VLM can reliably encode and expose causal structures; however, this assumption may not hold in domains characterized by highly specialized or rapidly evolving knowledge. Moreover, the identification of causal pathways during inference necessitates additional model queries, thereby increasing computational overhead and potentially constraining scalability in practical deployments.

\section{Conclusion}
We present \our, a multimodal causal retrieval framework that enhances factuality and robustness in medical vision–language models. By integrating graph-based causal reasoning within cross-modal retrieval, \our~achieves state-of-the-art results on radiology-specific VQA and report generation tasks. Ablation analysis further underscores the importance of causal grounding for clinically meaningful evidence and demonstrates the effectiveness of manual refinement in improving precision. Taken together, these findings highlight causal retrieval as a viable pathway toward safer deployment in real-world clinical settings.


\vfill\pagebreak


{
\normalsize
\bibliographystyle{IEEEbib}
\bibliography{refs_merged}   

@misc{ma_causal_2024,
	title = {Causal {Inference} with {Large} {Language} {Model}: {A} {Survey}},
	shorttitle = {Causal {Inference} with {Large} {Language} {Model}},
	url = {http://arxiv.org/abs/2409.09822},
	language = {en},
	urldate = {2024-09-30},
	publisher = {arXiv},
	author = {Ma, Jing},
	month = sep,
	year = {2024},
	note = {arXiv:2409.09822 [cs]},
}

@article{manzari2023medvit,
  title={MedViT: a robust vision transformer for generalized medical image classification},
  author={Manzari, Omid Nejati and Ahmadabadi, Hamid and Kashiani, Hossein and Shokouhi, Shahriar B and Ayatollahi, Ahmad},
  journal={Computers in biology and medicine},
  volume={157},
  pages={106791},
  year={2023},
  publisher={Elsevier}
}

@article{johnson2019mimic,
  title={MIMIC-CXR: A large publicly available database of labeled chest radiographs},
  author={Johnson, Alistair EW and others},
  journal={Scientific Data},
  year={2019}
}

@article{wu2023towards,
  title={Towards generalist foundation model for radiology by leveraging web-scale 2D\&3D medical data},
  author={Wu, Chaoyi and Zhang, Xiaoman and Zhang, Ya and Wang, Yanfeng and Xie, Weidi},
  journal={arXiv preprint arXiv:2308.02463},
  year={2023}
}

@article{li2023llava,
  title={Llava-med: Training a large language-and-vision assistant for biomedicine in one day},
  author={Li, Chunyuan and Wong, Cliff and Zhang, Sheng and Usuyama, Naoto and Liu, Haotian and Yang, Jianwei and Naumann, Tristan and Poon, Hoifung and Gao, Jianfeng},
  journal={Advances in Neural Information Processing Systems},
  volume={36},
  pages={28541--28564},
  year={2023}
}

@article{hu2022lora,
  title={Lora: Low-rank adaptation of large language models.},
  author={Hu, Edward J and Shen, Yelong and Wallis, Phillip and Allen-Zhu, Zeyuan and Li, Yuanzhi and Wang, Shean and Wang, Lu and Chen, Weizhu and others},
  journal={ICLR},
  volume={1},
  number={2},
  pages={3},
  year={2022}
}

@article{hu2021lora,
  title={Lora: Low-rank adaptation of large language models},
  author={Hu, Edward J and Shen, Yelong and Wallis, Phillip and Allen-Zhu, Zeyuan and Li, Yuanzhi and Wang, Shean and Wang, Lu and Chen, Weizhu},
  journal={arXiv},
  year={2021}
}

@article{iu-xray,
  title={Preparing a collection of radiology examinations for distribution and retrieval},
  author={Demner-Fushman, Dina and Kohli, Marc D and Rosenman, Marc B and Shooshan, Sonya E and Rodriguez, Laritza and Antani, Sameer and Thoma, George R and McDonald, Clement J},
  journal={JAMIA},
  year={2016},
}

@article{demner2016preparing,
  title={Preparing a collection of radiology examinations for distribution and retrieval},
  author={Demner-Fushman, Dina and Kohli, Marc D and Rosenman, Marc B and Shooshan, Sonya E and Rodriguez, Laritza and Antani, Sameer and Thoma, George R and McDonald, Clement J},
  journal={Journal of the American Medical Informatics Association},
  volume={23},
  number={2},
  pages={304--310},
  year={2016},
  publisher={Oxford University Press}
}

@article{gao2023retrieval,
  title={Retrieval-augmented generation for large language models: A survey},
  author={Gao, Yunfan and Xiong, Yun and Gao, Xinyu and Jia, Kangxiang and Pan, Jinliu and Bi, Yuxi and Dai, Yi and Sun, Jiawei and Wang, Haofen},
  journal={arXiv preprint arXiv:2312.10997},
  year={2023}
}

@article{leng2023mitigating,
  title={Mitigating object hallucinations in large vision-language models through visual contrastive decoding},
  author={Leng, Sicong and Zhang, Hang and Chen, Guanzheng and Li, Xin and Lu, Shijian and Miao, Chunyan and Bing, Lidong},
  journal={arXiv preprint arXiv:2311.16922},
  year={2023}
}

@article{huang2023opera,
  title={OPERA: Alleviating Hallucination in Multi-Modal Large Language Models via Over-Trust Penalty and Retrospection-Allocation},
  author={Huang, Qidong and Dong, Xiaoyi and Zhang, Pan and Wang, Bin and He, Conghui and Wang, Jiaqi and Lin, Dahua and Zhang, Weiming and Yu, Nenghai},
  journal={arXiv preprint arXiv:2311.17911},
  year={2023}
}

@article{he2024meddr,
  title={MedDr: Diagnosis-Guided Bootstrapping for Large-Scale Medical Vision-Language Learning},
  author={He, Sunan and Nie, Yuxiang and Chen, Zhixuan and Cai, Zhiyuan and Wang, Hongmei and Yang, Shu and Chen, Hao},
  journal={arXiv preprint arXiv:2404.15127},
  year={2024}
}

@article{alsentzer2019publicly,
  title   = {Publicly available clinical BERT embeddings},
  author  = {Alsentzer, Emily and Murphy, John R and Boag, Willie and Weng, Wei-Hung and Jin, Di and Naumann, Tristan and McDermott, Matthew},
  journal = {arXiv preprint arXiv:1904.03323},
  year    = {2019}
}

@article{chuang2023dola,
  title={Dola: Decoding by contrasting layers improves factuality in large language models},
  author={Chuang, Yung-Sung and Xie, Yujia and Luo, Hongyin and Kim, Yoon and Glass, James and He, Pengcheng},
  journal={arXiv preprint arXiv:2309.03883},
  year={2023}
}

@book{pearl2009causality,
  title={Causality: Models, Reasoning, and Inference},
  author={Pearl, Judea},
  year={2009},
  publisher={Cambridge university press}
}

@inproceedings{xiong2024benchmarking,
  title={Benchmarking retrieval-augmented generation for medicine},
  author={Xiong, Guangzhi and Jin, Qiao and Lu, Zhiyong and Zhang, Aidong},
  booktitle={Findings of the Association for Computational Linguistics ACL 2024},
  pages={6233--6251},
  year={2024}
}

@article{xia2024mmed,
  title={Mmed-rag: Versatile multimodal rag system for medical vision language models},
  author={Xia, Peng and Zhu, Kangyu and Li, Haoran and Wang, Tianze and Shi, Weijia and Wang, Sheng and Zhang, Linjun and Zou, James and Yao, Huaxiu},
  journal={arXiv preprint arXiv:2410.13085},
  year={2024}
}

@article{wang2025causalrag,
  title={Causalrag: Integrating causal graphs into retrieval-augmented generation},
  author={Wang, Nengbo and Han, Xiaotian and Singh, Jagdip and Ma, Jing and Chaudhary, Vipin},
  journal={arXiv preprint arXiv:2503.19878},
  year={2025}
}

@article{chen2025cross,
  title={Cross-Modal Causal Representation Learning for Radiology Report Generation},
  author={Chen, Weixing and Liu, Yang and Wang, Ce and Zhu, Jiarui and Li, Guanbin and Liu, Cheng-Lin and Lin, Liang},
  journal={IEEE Transactions on Image Processing},
  year={2025},
  publisher={IEEE}
}

@inproceedings{chen2024domain,
  title={Domain game: Disentangle anatomical feature for single domain generalized segmentation},
  author={Chen, Hao and Zhang, Hongrun and Chan, U Wang and Yin, Rui and Wang, Xiaofei and Li, Chao},
  booktitle={International Workshop on Computational Mathematics Modeling in Cancer Analysis},
  pages={41--51},
  year={2024},
  organization={Springer Nature Switzerland Cham}
}

@article{xia2024rule,
  title={RULE: Reliable Multimodal RAG for Factuality in Medical Vision Language Models},
  author={Xia, Peng and Zhu, Kangyu and Li, Haoran and Zhu, Hongtu and Li, Yun and Li, Gang and Zhang, Linjun and Yao, Huaxiu},
  journal={arXiv preprint arXiv:2407.05131},
  year={2024}
}

@article{sun2024fact,
  title={Fact-Aware Multimodal Retrieval Augmentation for Accurate Medical Radiology Report Generation},
  author={Sun, Liwen and Zhao, James and Han, Megan and Xiong, Chenyan},
  journal={arXiv preprint arXiv:2407.15268},
  year={2024}
}

@article{chen2024detecting,
  title={Detecting and Evaluating Medical Hallucinations in Large Vision Language Models},
  author={Chen, Jiawei and Yang, Dingkang and Wu, Tong and Jiang, Yue and Hou, Xiaolu and Li, Mingcheng and Wang, Shunli and Xiao, Dongling and Li, Ke and Zhang, Lihua},
  journal={arXiv preprint arXiv:2406.10185},
  year={2024}
}
}

\end{document}